\def\year{2021}\relax
\documentclass[letterpaper]{article} 
\usepackage{aaai21}  
\usepackage{times}  
\usepackage{helvet} 
\usepackage{courier}  
\usepackage[hyphens]{url}  
\usepackage{graphicx} 
\usepackage{graphicx}  
\usepackage{caption} 
\usepackage{amsmath}
\usepackage{algorithm}  
\usepackage{algorithmic} 
\usepackage{multirow}
\usepackage{amsfonts}
\usepackage{booktabs}
\usepackage[switch]{lineno}
\newcommand{\R}{\mathbb{R}}

\newcommand{\mG}{\mathcal{G}}
\newcommand{\BX}{\mathbf{X}}
\newcommand{\BZ}{\mathbf{Z}}

\title{Bayesian Spatio-Temporal Graph Convolutional Network for Traffic Forecasting}
\author {
        Jun Fu\textsuperscript{\rm 1}
        Wei Zhou, \textsuperscript{\rm 1}
        Zhibo Chen \textsuperscript{\rm 1} \\
}
\affiliations {
    \textsuperscript{\rm 1} University of Science and Technology of China \\
    fujun@mail.ustc.edu.cn, weichou@mail.ustc.edu.cn, chenzhibo@ustc.edu.cn
}
\begin{document}
	\maketitle
	\begin{abstract}
    In traffic forecasting, graph convolutional networks (GCNs), which model traffic flows as spatio-temporal graphs, have achieved remarkable performance. However, existing GCN-based methods heuristically define the graph structure as the physical topology of the road network, ignoring potential dependence of the graph structure over traffic data. And the defined graph structure is deterministic, which lacks investigation of uncertainty. In this paper, we propose a Bayesian Spatio-Temporal Graph Convolutional Network (BSTGCN) for traffic prediction. The graph structure in our network is learned from the physical topology of the road network and traffic data in an end-to-end manner, which discovers a more accurate description of the relationship among traffic flows. Moreover, a parametric generative model is proposed to represent the graph structure, which enhances the generalization capability of GCNs. We verify the effectiveness of our method on two real-world datasets, and the experimental results demonstrate that BSTGCN attains superior performance compared with state-of-the-art methods.
	\end{abstract}
	\section{Introduction}
	\par Traffic congestion is a growing drain on the economy with the acceleration of urbanization. For example, the cost of traffic congestion in America reached \$124 billion in 2014, and will rise to \$186 billion in 2030, according to a report by Forbes~\cite{guerrini2014traffic}. Therefore, improving traffic conditions is essential for increasing city efficiency, improving economy, and easing people’s daily life. One promising way to mitigate urban traffic congestion is to introduce Intelligent Transportation Systems (ITS), in which traffic prediction plays a vital role. However, accurate traffic prediction is still challenging due to the complex spatio-temporal dependence among traffic flows. 
	\par In the past few decades, many schemes have been proposed for traffic prediction, which can be broadly divided into two categories: temporal dependence based methods and spatio-temporal dependence based methods. Temporal dependence based methods leverage temporal characteristics of traffic flows to predict future traffic condition. Nevertheless, these methods have limited capability to achieve accurate traffic prediction due to ignoring the spatial dependence among traffic flows. Therefore, spatio-temporal dependence based methods are increasingly emerging, which take the spatial information into account. Since the road network is naturally structured as a graph in a non-Euclidean space with roads as nodes and their natural connections as edges, researchers prefer to use graph convolutional networks (GCNs)~\cite{kipf2016semi} to model spatial dependence among traffic flows instead of convolutional neural networks (CNNs). 
	\par However, existing GCN-based methods have two main disadvantages in terms of the graph construction: (1) The graph structure employed in GCNs is heuristically predefined and represents only the physical structure of the road network. Thereby, it is not guaranteed to be optimal description of dependence among traffic flows. For example, the relationship between two traffic flows that have similar trends but are located far away from each other is also important for traffic prediction. However, such dependence cannot be captured in the predefined road-topology-based graphs. (2) Introducing uncertainty into the graph structure, such as randomly dropping nodes or edges, could enhance the generalization capability of GCNs. Nevertheless, the graph structure employed in existing GCN-based methods is deterministic, which is lack of uncertainty investigation.
	\par To solve above issues, a Bayesian Spatio-Temporal Graph Convolutional Network (BSTGCN) is proposed in this paper. It views the graph structure as a sample drawn from a parametric generative model, and aims to infer the posterior probability of the graph structure based on two types of information. One type is the physical topology of the road network, the other type is traffic data. In addition, parameters of the generative model are optimized together with weights of GCNs by the back-propagation algorithm in an end-to-end manner. The main contributions of our work lie in three folds:
	\begin{itemize}
		\item This work proposes to learn the graph structure from the physical topology of the road network and traffic data in an end-to-end manner, which can find a more accurate descriptor of the relationship among traffic flows. 
		\item A generative model is proposed to represent the graph structure, which can improve the generalization capability of GCNs.
		\item We validate the effectiveness of our method on two real-world datasets, and the experimental results show that our approach outperforms state-of-the-art methods by a noticeable margin.
	\end{itemize}
	\par The rest of the paper is organized as follows. Section 2 reviews research works related to traffic prediction. Section 3, 4, and 5 introduce the background, details of our method, and experimental results, respectively. Section 6 concludes this paper and points out some future directions.
	
	\section{Related Work}
	Traffic prediction has attracted a lot of attention in recent years due to its essential role in traffic management. Current methods generally have two categories, temporal dependence based methods, and spatio-temporal dependence based methods. Temporal dependence based methods only consider the temporal characteristics of traffic flows for traffic prediction. At the early stage, auto-regressive models including Auto-Regressive Integrated Moving Average (ARIMA)~\cite{ahmed1979analysis}, Kalman filtering model~\cite{okutani1984dynamic}, and seasonal ARIMA~\cite{williams2003modeling} are widely used in traffic prediction. However, these statistical models rely on the stationary assumption on traffic time series data, which hinders their performance on real-world traffic conditions that vary over time. Some traditional machine learning methods including Linear SVR~\cite{wu2004travel}, and random forest regression~\cite{leshem2007traffic} are also tailored to solve traffic prediction, but are limited by hand-crafted features and shallow architectures. With the rapid development of deep learning, a variety of neural network architectures are applied in traffic prediction, such as the Feed Forward Neural network~\cite{park1999forecasting}, LSTM and GRU~\cite{fu2016using}. Despite their impressive capability in modeling temporal dynamics, these methods still have limited ability to achieve accurate traffic prediction due to lacking consideration of the spatial dependence among traffic flows. 
	\par Spatio-temporal dependence based methods take temporal nature and spatial dependence of traffic flows into account for traffic prediction. To capture the spatial dependence among traffic flows, early attempts including SAE~\cite{lv2014traffic}, ST-ResNet~\cite{zhang2017deep}, and SRCN~\cite{yu2017spatiotemporal} have tried to employ various CNNs. Nevertheless, considering that CNNs prefer to Euclidean data~\cite{defferrard2016convolutional}, such as images, regular grids, and so on, such methods can not perform well in the road network with complex topological structure. As a result, a temporal graph convolutional network (T-GCN)~\cite{zhao2019t} is proposed for traffic prediction, where GCNs and GRUs are combined to capture spatio-temporal features of traffic flows. Later, A3T-GCN~\cite{zhu2020a3t} boosts the performance of T-GCN through introducing an attention mechanism. However, existing GCN-based works ignore the uncertainty and the information of traffic flows in the process of graph construction.

	
	\section{Background}
	\subsection{Graph Convolutional Network}
	\par GCNs have been widely applied to a broad of applications, such as semi-supervised learning~\cite{jiang2019semi}, action recognition~\cite{yan2018spatial}, and quality assessment~\cite{xu2020blind}. Graph convolutional operation can be designed in either spatial or spectral domain. In this paper, we focus on the latter. Spectral convolution on graph $\mathcal{G}$ with $N$ nodes is defined as the product of a signal $\BX \in \text{R}^{N\times n}$  (each node with a $n$-dim feature) and a filter $g_\theta(L)$ parameterized by $\theta$, i.e.:
	\begin{equation}
	g_\theta(L) * \BX = U g_\theta U^T\BX,
	\label{eq:gnns}
	\end{equation}
	where $U$ is the eigenvector of normalized Laplacian matrix $L$, and $U^T\BX$ is the Fourier transform of $\BX$. Since the Fourier transform is computationally expensive, a faster propagation rule~\cite{kipf2016semi} is proposed, i.e.:
	\begin{equation}
	H^{(l)} = \sigma(\widehat{A}H^{(l-1)}W^{(l)}),
	\label{eq:gnns1}
	\end{equation}
	where $H^{(l)}$ and $W^{(l)}$ are the output feature and trainable parameters of the layer $l$, $\sigma$ denotes the activation function, and $H^{(0)}$ equals $\BX$. The normalized adjacency matrix $\widehat{A}$ is equal to ${D}^{-\frac{1}{2}}\,{A} \,{D}^{-\frac{1}{2}}$, where $A$ and $D$ are the adjacency matrix and the degree matrix of $\mathcal{G}$. To include self-loop information, Eq.~\ref{eq:gnns1} is rewritten as:
	\begin{equation}
	H^{(l)} = \sigma(\widetilde{A}H^{(l-1)}W^{(l)}),
	\label{eq:gnns2}
	\end{equation}
	where $\widetilde{A} = \widehat{A} + I_N$, and $I_N$ is the identity matrix.
	\subsection{Bayesian Graph Convolutional Network}
		\begin{figure*}[htbp]
		\centering
		\includegraphics[width=1.0\linewidth]{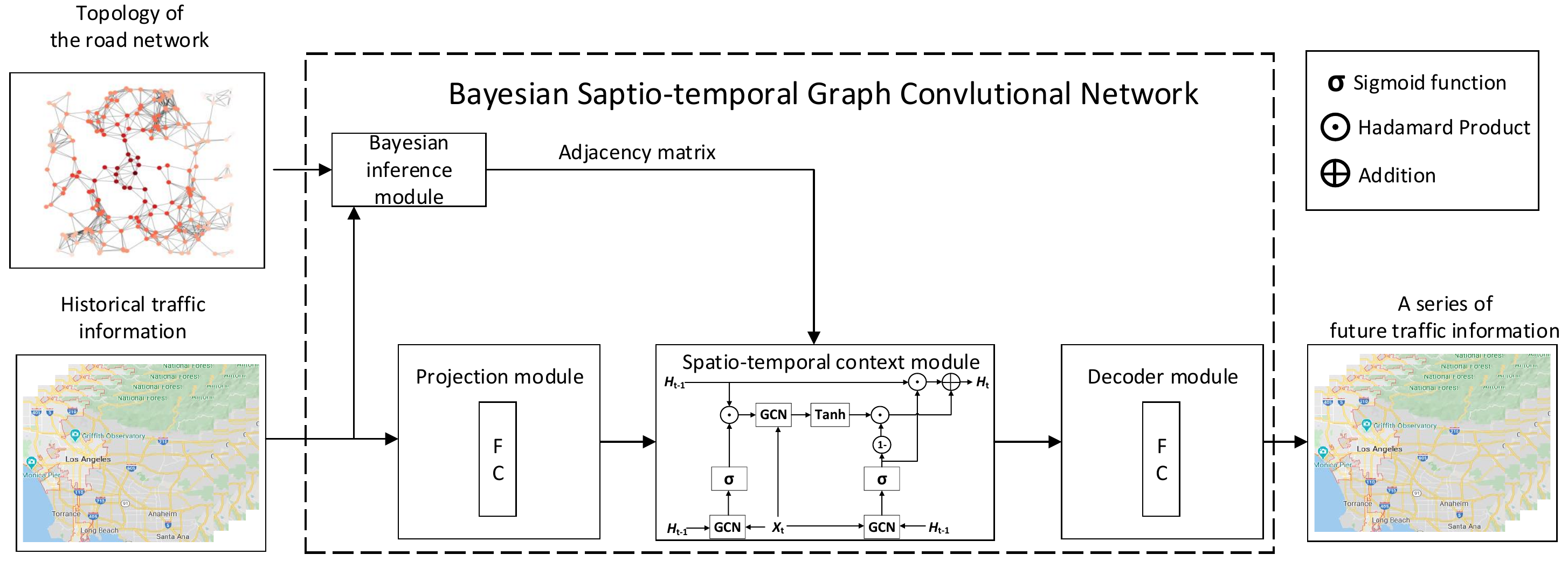}
		\caption{The illustration of the proposed Bayesian spatio-temporal graph convolutional network. FC and GCN denote the fully-connected layer and graph convolutional network.}
		\label{fig:BGCN}
	\end{figure*}
	BGCN~\cite{zhang2019bayesian} is firstly introduced in the task of semi-supervised node classification. In this task, we have an observed graph $\mathcal{G}_{obs} = (\mathcal{V},\mathcal{E})$, where $\mathcal{V}$ and $\mathcal{E}$ denote the set of $N$ nodes and edges. And we obtain feature vectors of all nodes $\BX = \{x_1,...,x_N\}$, but only know labels of a portion of nodes $\mathbf{Y_{\mathcal{L}}} = \{y_1,...,y_n\}$. Then, we aim to infer the labels of the remaining nodes based on $\mathcal{G}_{obs}$, $\BX$ and $\mathbf{Y_{\mathcal{L}}}$. In the BGCN based framework viewing the graph structure and the weights of GCNs as random variables, the goal is to infer the posterior probability of labels $\BZ$, i.e.:
    \begin{align}
    p(\BZ|\mathbf{Y_{\mathcal{L}}},\BX,\mG_{obs}) &= \int p(\BZ|W,\mG,\BX) p(W|\mathbf{Y_{\mathcal{L}}},\BX,\mG)\,\nonumber\\
    & \quad \quad p(\mG|\lambda) p(\lambda|\mG_{obs}) \,dW\,d\mG\,d\lambda\label{eq:margin1}\,,
    \end{align}
	where $\lambda$ parameterizes random graphs and the term $p(\BZ|W,\mG,\BX)$ is modeled by a categorical distribution. However, this formula ignores any possible dependence of the graph $\mG$ on data as it targets the inference of $p(\mG|\mG_{obs})$. To this end, an enhanced BGCN~\cite{pal2019bayesian} is proposed, and formulates the posterior predictive distribution as follows:
    \begin{align}
	p(\BZ|\mathbf{Y_{\mathcal{L}}},\BX,\mG_{obs}) &= \int p(\BZ|W,\mG,\BX)p(W|\mathbf{Y_{\mathcal{L}}},\BX,\mG)\,\nonumber \\
	& \quad \quad p(\mG|\mG_{obs},\BX,\mathbf{Y_{\mathcal{L}}})dWd\mG,\label{eq:margin2}
    \end{align}
	where the item $p(\mG|\mG_{obs},\BX,\mathbf{Y_{\mathcal{L}}})$ allows us to introduce the information of data. Additionally, as the integral in Eq.~\ref{eq:margin2} is intractable, a Monte Carlo approximation~\cite{gal2016dropout} is often involved as follows: 
	\begin{align}
		p(\BZ|\mathbf{Y_{\mathcal{L}}},\BX,\mG_{obs}) \approx 
		\frac{1}{SC} \sum_{s=1}^S
		\sum_{c=1}^{C} p(\BZ|W_{s,c},\mG_{obs},\BX),
		\label{eq:MC_posterior1}
	\end{align}
	where $C$ graphs $\mG_{c}$ sampled from $p(\mG|\mG_{obs},\BX,\mathbf{Y_{\mathcal{L}}})$ and $S$ weights samples $W_{s,c}$ drawn from $p(W|\mathbf{Y_{\mathcal{L}}},\BX,\mG)$. Considering that sampling graph from the posterior probability of $\mG$ is time consuming, the maximum a posterior (MAP) inference of $\mG$ is introduced as follows:
	\begin{align}
		\hat{\mG} &= \arg\max_{\mG} p(\mG|\mG_{obs}, \mathbf{X}, \mathbf{Y_{\mathcal{L}}}).
		\label{opt:graph_inference}
	\end{align}
	As a result,  Eq.~\ref{eq:MC_posterior1} can be simplified as follows:
	\begin{align}
			p(\BZ|\mathbf{Y_{\mathcal{L}}},\BX,\mG_{obs})  \approx \frac{1}{S}\sum_{s=1}^S p(\BZ|W_{s},\hat{\mG},\BX),
	\label{eq:approx_posterior_map}
	\end{align}
	where $S$ weights samples $W_s$ obtained via dropout. 
	
	\section{Method}
	\subsection{Problem Formulation}
	\par In the task of traffic prediction, we have access to the topology of the road network $\mathcal{G}_{obs}=(\mathcal{V},\mathcal{E})$, where $\mathcal{V}$ is the set of ${N}$ roads and $\mathcal{E}$ denotes the set of edges. And we also obtain historical traffic data on roads $X_{1:t} = \{X_1, X_2,...,X_t\}$, where $X_t$ belongs to $\mathbb{R}^{N \times n}$, and $n$ is the dimension of traffic data. Then, our goal is to forecast $T$ future signals $X_{t+1:t+T}$ based on $\mathcal{G}_{obs}$ and $X_{1:t}$ in an iterative manner. In other words, to predict $X_{t+T}$, we take observed signals $X_{1:t}$ and previous estimated results $X_{t+1:t+T-1}$ as the input. It is worth noting that the traffic data on roads only includes the traffic speed in this paper, i.e., $n$ is set to 1.
	
	\par In general, traffic prediction is regarded as a task of learning a non-linear mapping function $F$ which maps historical traffic data and the topology of the road network into traffic conditions in the future timestamps. Mathematically, we formulate the objective of traffic prediction as follows:
	\begin{equation}
	\mathop{\arg\min}_{F} \sum_{i=1}^{T} \| Y_{t+i} - F(X_{1:t+i-1}, \mathcal{G}_{obs})\|^2, 
	\label{eq:goal}
	\end{equation}
	where $Y_{t+i}$ is the ground truth at time $t+i$, and $i$ is the desirable horizon ahead of the current timestamp $t$. In this paper, we propose a novel Bayesian spatio-temporal graph convolutional network (BSTGCN) to model $F$. As shown in Figure~\ref{fig:BGCN}, four modules are involved in the BSTGCN, including a projection module, a spatio-temporal context module, a decoder module, and a Bayesian inference module. Next, we will detail these modules in sequence.
	\subsection{Projection Module}
	The projection module is designed to project 1D traffic speed of each road into a high-dimensional feature space using a linear layer. Concretely, the traffic speed of $N$ roads at time $t$, $X_t \in \R^{N\times1}$, is transformed into 64-dim traffic features $L_t \in \R^{N\times64}$ as follows: 
	\begin{equation}
	L_t = f(X_t), 
	\end{equation}
	where the function $f(\cdot)$ consists of a fully-connected (FC) layer with 64 neurons, and the enhanced representation of traffic speed $L_t$ is beneficial to learn the spatio-temporal characteristics of traffic flows.
	\subsection{Spatio-temporal Context Module}	
	As aforementioned, the future traffic speed of one road not only depends on the historical traffic speed of the road, but also is constrained to the physical topology of the road network. As a result, the spatio-temporal context module is proposed to capture the spatio-temporal features for each road. In the light of the good performance of Gate Recurrent Unit (GRU)~\cite{cho2014properties} in modeling temporal characteristics of sequential data~\cite{lu2019gru}, we tailor the original GRU for dealing with graph-structured traffic data. In particular, we sequentially feed the input sequence $L_{1:t}$ into the tailored GRU, and denote the output of the tailored GRU at timestamp $t+1$ as: %
	\begin{equation}
	 H_{t+1} = h(L_1,...,L_{t}),
	\end{equation}
	where the function $h(\cdot)$ is comprised of a tailored GRU with the hidden size of 64, and the input-state transitions of the tailored GRU are expressed as follows:
	\begin{equation}
    	\begin{split}
    	u_{t+1} &=\sigma(W_{u} \star \left[L_{t},H_{t}\right]+b_{u})  \\
    	r_{t+1} &= \sigma(W_{r} \star \left[L_{t},H_{t}\right]+b_{r}) \\
    	n_{t+1} &=\tanh(W_{c} \star \left[L_{t},r_{t+1}\odot H_{t}\right]+b_{c}) \\
    	H_{t+1} &=u_{t}\odot H_{t}+(1-u_{t+1})\odot n_{t+1},
    	\label{eq:GRU}
    	\end{split}
	\end{equation}
    where $H_{t+1}$, $r_{t+1}$, $z_{t+1}$, and $n_{t+1}$ are the hidden state, the reset, update, and new gates at time $t+1$, respectively. $\sigma$ is the sigmoid function, $\star$ denotes the graph convolutional operation, and $\odot$ means the Hadamard product. $W_{u}$, $W_{r}$, $W_{c}$, $b_u$, $b_r$, and $b_c$ are trainable parameters. As seen in Eq.~\ref{eq:GRU}, the tailored GRU replaces FC layers in the conventional GRU with GCNs, which could learn the spatial characteristics of traffic flows. Moreover, we introduce uncertainty and the information of traffic data to the graph structure employed in GCNs, which is detailed in the part of the Bayesian inference module.
	\subsection{Decoder Module}	
    \par The decoder module aims to map the spatio-temporal feature $H_{t+1}$ into the traffic speed of roads at time $t+1$ with a linear layer, which is formulated as follows:
	\begin{equation}
	    X_{t+1} = z(H_{t+1}),
	\end{equation}
	where the function $z(\cdot)$ is made up of a FC layer with 1 neuron. 
	\subsection{Bayesian Inference Module}		
	\par The Bayesian inference module is designed to discover better graph structure from the $\mathcal{G}_{obs}$, input data $\mathbf{X}=X_{1:t}$, and corresponding label $\mathbf{Y}=Y_{t+1}$, as well as introducing uncertainty into the graph structure. In this paper, we consider a Bayesian approach, viewing the graph structure as a sample drawn from a parametric generative model. We then aim to infer the posterior predictive distribution as follows:
	\begin{equation}
	\begin{split}
	p(\mathbf{Z}|\mathbf{X},\mathbf{Y},\mathcal{G}_{obs},W) &= \int p(\mathbf{Z}|\mathbf{X},W,\mathcal{G}) p(\mathcal{G}|\mathbf{X},\mathbf{Y},\mathcal{G}_{obs})d\mathcal{G} \\
	&=\int p(\mathbf{Z}|\mathbf{X},W,\mathcal{G})p(\mathcal{G}|g,\mathbf{X},\mathbf{Y})\\
	& \qquad\quad p(g|\mathcal{G}_{obs})d\mathcal{G}dg,
	\end{split}
	\label{bgcn:integral}
	\end{equation}
	where $W$ is the set of trainable parameters of our network, and $\mathbf{Z}=X_{t+1}$ is the ground-truth traffic speed at time $t+1$. As seen in Eq.~\ref{bgcn:integral}, the posterior probability of graph $p(\mathcal{G}|\mathbf{X},\mathbf{Y},\mathcal{G}_{obs})$ is calculated in a two-step manner, where the information of the topology of the road network $\mathcal{G}_{obs}$ and the information of traffic data $(\mathbf{X},\mathbf{Y})$ are successively introduced into the graph structure $\mathcal{G}$. 
	
	\begin{algorithm}[tb]
		\caption{Training methodology of BSTGCN}
		\label{alg:bgcn_non_param}
		\begin{algorithmic}[1] 
			\STATE \textbf{ Input:} $\mathcal{G}_{obs}$, training datasets $D$
			\STATE \textbf{ Output:} $W$, $A_{\hat{g}}$, $\phi$
			\STATE randomly initialize $W$ and $\phi$
			\STATE obtain $A_{\hat{g}}$ via solving Eq.~\ref{opt:adj_inference}
			\FOR{ $epoch=1$ {\bfseries to} $100$}
			\STATE sample the graph structure $\mathcal{G}$ from $A_{\hat{g}} + \phi$ via dropout
			\STATE sample a batch of data $\{X_{1:t},Y_{t+1:t+T}\}$ from $D$
			\FOR{$i=1$ {\bfseries to} $T$}
			\STATE obtain predicted result $\mathbf{X_{t+i}}$ at time $i$
			\ENDFOR
			\STATE optimize $W$ and $\phi$ by minimizing Eq.~\ref{eq:goal}  
			\ENDFOR
		\end{algorithmic}
	\end{algorithm}	
	
    \par The specific calculation of Eq.~\ref{bgcn:integral} is presented as follows. Since there is no closed-form solution for the integral in Eq.~\ref{bgcn:integral}, a Monte Carlo approximation is introduced as follows:
	\begin{align}
		p(\BZ|\mathbf{Y},\BX,\mG_{obs}) \approx 
		\frac{1}{SC} \sum_{s=1}^S
		\sum_{c=1}^{C} p(\BZ|W,\mG_{s,c},\BX),
		\label{eq:MC_posterior}
	\end{align}    
	where $C$ graphs $g_{c}$ sampled from $p(g|\mG_{obs})$, $S$ weights samples $\mG_{s,c}$ drawn from $p(\mathcal{G}|g_c,\mathbf{X},\mathbf{Y})$, and $p(\mathbf{Z}|\mathbf{X},W,\mG_{s,c})$ is modeled by a Gaussian likelihood. Like the improved BGCN~\cite{pal2019bayesian},  we replace the integral over $g$ with a MAP process, as follows:  
	\begin{equation}
	    \hat{g} = \mathop{\arg\max}_{g} \ \ p(g|\mathcal{G}_{obs}).
	\label{bgcn:map}
	\end{equation}
	As described in the work~\cite{pal2019bayesian}, solving Eq.~\ref{bgcn:map} is equivalent to learning a $N \times N$ symmetric adjacency matrix of $g$, expressed as follows:
	\begin{equation}
	\begin{split}
	A_{\hat{g}}= \mathop{\arg\min}_{\substack{A_{g} \in \mathbf{R_+}^{N\times N}, \\ A_{g}=A_{g}^T }} \|A_{g} \odot Z\|_1 -\alpha\mathbf{1}^T\log(A_{g}\mathbf{1})  + \beta\|A_{g}\|^2, 
	\label{opt:adj_inference}
	\end{split}
	\end{equation}
	where $\alpha$ and $\beta$ control the scale and density of $A_{\hat{g}}$. Here, $Z$ is the pairwise distance of roads in the embedding space, which is calculated as follows:
	\begin{equation}
	Z_{p,q} = \|e_p - e_q \|^2,
	\end{equation}
	where $e_p$ and $e_q$ are the embedding vector of the $p$-th and $q$-th road. In this paper, we learn embedding vectors of roads through the Graph Variational Auto-Encoder algorithm~\cite{kipf2016variational}. After obtaining $Z$, we solve the Eq.~\ref{opt:adj_inference} via the prevalent method~\cite{kalofolias2017large}. As for the inference of $p(\mathcal{G}|g,\mathbf{X},\mathbf{Y})$, we also adopt a Monte Carlo approximation. Thus, Eq.~\ref{bgcn:integral} is rewritten as follows:
	\begin{equation}
	    p(\mathbf{Z}|\mathbf{X},\mathbf{Y},\mathcal{G}_{obs},W) \approx \frac{1}{S} \sum_{s=1}^{S} p(\mathbf{Z}|\mathbf{X},W,\mG_s),
	\label{bgcn:mc}
	\end{equation} 
	where $S$ weights samples $\mG_s$ drawn from $A_{\hat{g}} + \phi$ via dropout, and $\phi \in \R^{N\times N}$ is a trainable deterministic variable that allows us to introduce the information of traffic data into the graph structure. As we can see, $\phi$ aims to learn the global graph structure as it is shared across all timestamps. The training algorithm of $\phi$ is described in Algorithm \ref{alg:bgcn_non_param}.
	\subsection{Relationship with Existing BGCNs}
	\par The improved BGCN~\cite{pal2019bayesian} cannot be applied in traffic prediction due to two main reasons: (1) To include the information of traffic data in the graph structure, it needs to pre-calculate the MAP result $\hat{\mathcal{G}}$ for each timestamp, which is time-consuming. (2) It relies on the assumption that the graph structure is symmetric in the calculation of $\hat{\mathcal{G}}$. However, this hypothesis is easily violated in the task of traffic prediction because the mutual influence of two traffic flows is usually unequal. In our proposed method, we learn the dependence of the graph structure over traffic data by injecting a trainable parameter $\phi$, without introducing extra calculation. Moreover, the learned graph structure $A_{\hat{g}} + \phi$ can be either symmetric or asymmetric.

	\section{Experiments}
	\subsection{Datasets and Evaluation Metrics}
	\par We verify our model on two real-world traffic datasets, SZ-taxi and Los-loop. The SZ-taxi dataset records the traffic speed of 156 major roads in Luohu District, Shen Zhen from Jan. 1 to Jan. 31, 2015. And the Los-loop dataset collects the traffic speed of 207 highways in Los Angeles County from Mar. 1 to Mar. 7, 2012. The traffic data is aggregated in the SZ-taxi and Los-loop datasets every 15 minutes and every 5 minutes. The topology of the road network is available in both datasets. In our experiment, we split both datasets into the training set and the evaluation set in a ratio of 4 and 1, and use observed 60-minute traffic speed to predict traffic conditions in the next 15, 30, 45, and 60 minutes. 
	\par We compare BSTGCN with the following state-of-the-art methods including historical average model (HA)~\cite{liu2004summary}, ARIMA~\cite{ahmed1979analysis}, SVR~\cite{smola2004tutorial}, GCN model~\cite{kipf2016semi}, GRU model~\cite{cho2014properties}, T-GCN model~\cite{zhao2019t}, and A3T-GCN model~\cite{zhu2020a3t} in terms of Root Mean Squared Error (RMSE), Mean Absolute Error (MAE), Accuracy (ACC), Coefficient of Determination ($\text{R}^{2}$), and Explained Variance Score (VAR). Higher ACC, $\text{R}^{2}$ and VAR values as well as lower RMSE and MAE values represent better prediction performance.
	\subsection{Implement Details}
	\par We implement BSTGCN on the PyTorch framework~\cite{paSZke2017automatic}. All trainable variables in BSTGCN are optimized by the Adam~\cite{kingma2014adam} optimizer. We train BSTGCN on a NVIDIA Geforce GTX 1080Ti GPU for 100 epochs. The learning rate is initialized as $10^{-2}$ and $10^{-3}$ for the SZ-taxi and Los-loop dataset, respectively. During training, 32 pairs are randomly generated from the training dataset per iteration, and the learning rate is decayed by 0.2 every 25 epochs. It is worth noting that we use the learned graph structure without dropout in the evaluation phase. 
	\subsection{Parameter Experiment and Ablation Study}
    \par First, we investigate the impact of hidden size and Monte Carlo dropout probability on the prediction performance in Table ~\ref{tb:hidden size} and ~\ref{tb:dropout}. As we can see, the prediction performance with 64 hidden neurons achieves the optimum on two real-world datasets. And, the optimal configurations of Monte Carlo dropout probability are 0.1 and 0.5 for the SZ-taxi and Los-loop dataset. As a result, we fix these hyperparameters in the following experiment. 
    \par Second, we verify the importance of introducing uncertainty and different ways of learning the graph structure in Table~\ref{table:ablation}. $B_c$ and $B_d$ are two variants of BSTGCN. $B_a$ and $B_b$ learn the individual traffic pattern for each timestamp through Graph Attention Network~\cite{velivckovic2017graph} and Self-Attention Network~\cite{vaswani2017attention}, but ignore the uncertainty. According to results of $B_c$, $B_d$, and BSTGCN, we can conclude that introducing uncertainty and the information of traffic data into the graph structure can steadily improve the performance of traffic prediction. In addition, we can see that $B_c$ that learns the global traffic pattern across all timestamps exceeds $B_a$ and $B_b$ by a clear margin, which confirms the advantage of our way to learn graph structure from traffic data.
	\begin{table}[htbp]
		\footnotesize
		\caption{The impact of hidden size on the prediction performance.}
		\centering
		\resizebox{80mm}{18mm}{
			\renewcommand{\arraystretch}{1}
			\begin{tabular}{c|c|c|ccccc}
				\hline
				\multirow{2}{*}{Dataset}&
				\multirow{2}{*}{T}&
				\multirow{2}{*}{Metric}&
				\multicolumn{5}{c}{Hidden size} \\
				\cline{4-8}
				&&&8&16&32&64&128\\
				\hline\hline
				\multirow{5}{*}{SZ-taxi}&
				\multirow{5}{*}{60min} 
				&RMSE&      4.0630 &  4.0490&  4.0510&  \textbf{4.0270}&  4.1090\\
				&&MAE&      2.7220 &  2.7000&   \textbf{2.6870}& 2.7010&  2.7270\\
				&&ACC&  0.7170 &  0.7179& 0.7178& \textbf{0.7195}& 0.7137\\
				&&R$^{2}$&     0.8446& 0.8496& 0.8495& \textbf{0.8513}& 0.8451\\
				&&VAR&       0.8489 & 0.8499& 0.8499& \textbf{0.8515}& 0.8451\\
				\hline
				\multirow{5}{*}{Los-loop}&
				\multirow{5}{*}{60min} 
				&RMSE&   10.810 &     7.7330        &7.5140      & \textbf{7.0840}           &7.0980 \\
				&&MAE&   6.9610 &      4.7280        &4.4720       &\textbf{4.1350}         &4.2290 \\
				&&ACC&  0.8156 & 0.8682      &0.8720      &\textbf{0.8793}          &0.8790 \\
				&&R$^{2}$&     0.3992 & 0.6924        &0.7096     &\textbf{0.7418 }         &0.7409 \\
				&&VAR&       0.4053 & 0.694       &0.7096    &\textbf{0.7418}          &0.7414 \\
				\hline
		\end{tabular}}
		\label{tb:hidden size}
	\end{table}
	\begin{table}[htbp]
		\footnotesize
		\caption{The impact of Monte Carlo dropout probability on the prediction performance.}
		\centering
		\resizebox{80mm}{18mm}{
			\renewcommand{\arraystretch}{1}
			\begin{tabular}{c|c|c|ccccc}
				\hline
				\multirow{2}{*}{Dataset}&
				\multirow{2}{*}{T}&
				\multirow{2}{*}{Metric}&
				\multicolumn{5}{c}{Monte Carlo dropout probability} \\
				\cline{4-8}
				&&&0.1&0.3&0.5&0.7&0.9\\
				\hline\hline
				\multirow{5}{*}{SZ-taxi}&
				\multirow{5}{*}{60min} 
				&RMSE&        \textbf{4.0270}&  4.0350&  4.0470  &  4.1020& 4.6620\\
				&&MAE&        2.6860 & \textbf{2.6720} &  2.7050  &  2.7290& 3.1970\\
				&&ACC&   \textbf{0.7195}& 0.7189& 0.7181 & 0.7143& 0.6752\\
				&&R$^{2}$&      \textbf{0.8513}& 0.8507& 0.8498 & 0.8457& 0.8006\\
				&&VAR&        \textbf{0.8514}& 0.8507&  0.85 & 0.8457& 0.8008\\
				\hline
				\multirow{5}{*}{Los-loop}&
				\multirow{5}{*}{60min} 
				&RMSE&        6.9940        & 6.8370         & \textbf{6.7330}            &\text{6.7760} & 6.9580\\
				&&MAE&        4.0460        & 3.9340        & \textbf{3.9180}     &3.9740 & 4.1390\\
				&&ACC&   0.8808      & 0.8835           & \textbf{0.8853}             &\text{0.8845} & 0.8814\\
				&&R$^{2}$&      0.7483        &0.7596       & \textbf{0.7668}          &\text{0.7638} & 0.7509\\
				&&VAR&        0.7485       &0.7601        & \textbf{0.7669}           &\text{0.7638} & 0.7511\\
				\hline
		\end{tabular}}
		\label{tb:dropout}
	\end{table}
    \begin{table}[htbp]
    \caption{Ablation study of the proposed BSTGCN on the Los-loop dataset with the prediction horizon of 60 minutes.}
    \resizebox{\linewidth}{!}{%
    \begin{tabular}{@{}lccccc@{}}
    \toprule
    Methods & \multicolumn{1}{|c}{$B_a$} & $B_b$ & $B_c$ & $B_d$ & BSTGCN  \\ \midrule
    Topology of the road network $\mathcal{G}_{obs}$ & \multicolumn{1}{|c}{\checkmark} &   &  &  &  \\
    MAP result $A_{\hat{g}}$ of $\mathcal{G}_{obs}$  & \multicolumn{1}{|c}{} &  & \checkmark & \checkmark & \checkmark \\
    Dependence of the graph structure over traffic data $\phi$ & \multicolumn{1}{|c}{} &   &  & \checkmark  & \checkmark \\
    Uncertainty & \multicolumn{1}{|c}{}  &  &  &  & \checkmark \\
    Self Attention Network & \multicolumn{1}{|c}{} & \checkmark &    &  & \\
    Graph Attention Network & \multicolumn{1}{|c}{\checkmark} &    &  &  & \\ \midrule
    RMSE & \multicolumn{1}{|c}{7.6700} & 7.4870 & 7.9860 & 7.0840  & {\bf 6.7330} \\
    MAE & \multicolumn{1}{|c}{4.4410} & 4.3250 & 4.6660 & 4.1350 & {\bf 3.9180} \\ 
    ACC & \multicolumn{1}{|c}{0.8693} & 0.8724 & 0.8639 & 0.8793  & {\bf 0.8853} \\
    R$^{2}$ &\multicolumn{1}{|c}{0.6973} & 0.7117 & 0.6719  & 0.7418 & {\bf 0.7668} \\     
    VAR & \multicolumn{1}{|c}{0.6976} & 0.7118 & 0.6721  & 0.7418 & {\bf 0.7669} \\ \bottomrule        
    \end{tabular}%
    }
    \label{table:ablation}
    \vspace{-2mm}
    \end{table}

	\begin{table*}
		\caption{The prediction results of the BSTGCN model and other baseline methods on the SZ-taxi and Los-loop dataset.}
		\centering
		\resizebox{180mm}{40mm}{
			\renewcommand{\arraystretch}{1.3}
			\begin{tabular}{c|c|cccccccc|cccccccc}
				\hline
				\multirow{2}{*}{T}&
				\multirow{2}{*}{Metric}&
				\multicolumn{8}{c|}{SZ-taxi}&
				\multicolumn{8}{c}{Los-loop} \\
				\cline{3-18}
				&&HA&ARIMA&SVR&GCN&GRU&T-GCN&A3T-GCN&BSTGCN&HA&ARIMA&SVR&GCN&GRU&T-GCN&A3T-GCN&BSTGCN\\
				\hline\hline
				\multirow{5}*{15min}
				&RMSE&4.2951&7.2406&4.1455&5.6596&3.9994& 3.9325 & \textbf{3.8989}& 3.9670&7.0970&10.044&6.0084&7.7922&5.2182&5.1264&{5.0904}&\textbf{4.7585}\\
				&MAE&2.7815&4.9824&2.6233&4.2367&\textbf{2.5955}&2.7145&2.6840&{2.6490}&3.7585&7.6832&3.7285&5.3525&{3.0602}&3.1802&3.1365&\textbf{2.9150}\\
				&ACC&0.7008&0.4463&0.7112&0.6107&0.7249&0.7295&\textbf{0.7318}&0.7237 &0.8792&0.8275&	0.8977&0.8673&0.9109&0.9127&{0.9133}&\textbf{0.9185}\\
				&R$^{2}$&0.8307&$\ast$&0.8423&0.6654&0.8329&0.8539&{0.8512}&\textbf{0.8557} &0.7382&0.0025&0.8123&0.6843&0.8576&0.8634&{0.8653}&\textbf{0.8810}\\
				&VAR&0.8307&0.0035&0.8424&0.6655&0.8329&0.8539&{0.8512}&\textbf{0.8557}&0.7382&$\ast$&0.8146&0.6844&0.8577&0.8634&{0.8653}&\textbf{0.8811}\\
				\hline
				\multirow{5}*{30min}
				&RMSE&4.3481&6.7899&4.1628&5.6918&4.0942&3.9740&\textbf{3.9228}&4.0010&7.9717&9.3450&6.9588&8.3353&6.2802&6.0598&{5.9974}&\textbf{5.6380}\\
				&MAE&2.8171&4.6765&{2.6875}&4.2647&{2.6906}&2.7522&2.7038&\textbf{2.6530}&4.1692&7.6891&3.7248&5.6118&{3.6505}&3.7466&3.6610&\textbf{3.3580}\\
				&ACC&0.6971&0.3845&0.7100&0.6085&0.7184&0.7267&\textbf{0.7302}&0.7213&0.8642&0.8275&0.8815&0.8581&0.8931&0.8968&{0.8979}&\textbf{0.9040}\\
				&R$^{2}$&0.8266&$\ast$&0.8410&0.6616&0.8249&0.8451&{0.8493}&\textbf{0.8531}&0.6709&0.0031&0.7492&0.6402&0.7957&0.8098&{0.8137}&\textbf{0.8354}\\
				&VAR&0.8266&0.0081&0.8413&0.6617&0.8250&0.8451&{0.8493}&\textbf{0.8532}&0.6709&$\ast$&0.7523&0.6404&0.7958&0.8100&{0.8137}&\textbf{0.8354}\\
				\hline
				\multirow{5}*{45min}
				&RMSE&4.3910&6.7852&4.1885&5.7142&4.1534&3.9910&\textbf{3.9461}&4.0110&8.7643&10.051&7.7504&8.8036&7.0343&6.7065&{6.6840}&\textbf{6.2130}\\
				&MAE&2.8480&4.6734&{2.7359}&4.2844&2.7743&2.7645&2.7261&\textbf{2.6780}&4.5646&7.6924&4.1288&5.9534&{4.0915}&4.1158&4.1712&\textbf{3.6440}\\
				&ACC&0.6941&0.3847&0.7082&0.6069&0.7143&0.7255&\textbf{0.7286}&0.7206&0.8507&0.8273&0.8680&0.8500&0.8801&0.8857&{0.8861}&\textbf{0.8942}\\
				&R$^{2}$&0.8232&$\ast$&0.8391&0.6589&0.8198&0.8436&{0.8474}&\textbf{0.8525}&0.6035&$\ast$&0.6899&0.5999&0.7446&0.7679&{0.7694}&\textbf{0.8008}\\
				&VAR&0.8232&0.0087&0.8397&0.6590&0.8199&0.8436&{0.8474}&\textbf{0.8526}&0.6036&0.0035&0.6947&0.6001&0.7451&0.7684&{0.7705}&\textbf{0.8008}\\	
				\hline
				\multirow{5}*{60min}
				&RMSE&4.4312&6.7708&4.2156&5.7361&4.0747&4.0099&\textbf{3.9707}&4.0270&9.4970&10.054&8.4388&9.2657&7.6621&7.2677&{7.0990}&\textbf{6.7330}\\
				&MAE&2.8754&4.6655&2.7751&4.3034&{2.7712}&2.7860&2.7391&\textbf{2.6860}&4.9491&7.6952&{4.5036}&6.2892&4.5186&4.6021&4.2343&\textbf{3.9180}\\
				&ACC&0.6913&0.3851&0.7063&0.6054&0.7197&0.7242&\textbf{0.7269}&0.7195&0.8382&0.8273&0.8562&0.8421&0.8694&0.8762&{0.8790}&\textbf{0.8853}\\
				&R$^{2}$&0.8199&$\ast$&0.8370&0.6564&0.8266&0.8421&{0.8454}&\textbf{0.8513}&0.5360&$\ast$&0.6336&0.5583&0.6980&0.7283&{0.7407}&\textbf{0.7668}\\
				&VAR&0.8199&0.0111&0.8379&0.6564&0.8267&0.8421&{0.8454}&\textbf{0.8514}&0.5361&0.0036&0.5593&0.5593&0.6984&0.7290&{0.7415}&\textbf{0.7669}\\
				\hline
		\end{tabular}}
		\label{tb:cmp}
	\end{table*}
	
	
	
	\subsection{Comparison with Existing Methods}
	Table~\ref{tb:cmp} summaries the comparison between BSTGCN and existing methods in terms of RMSE, MAE, ACC, $\text{R}^2$ and VAR under various prediction horizon of traffic prediction. The major findings are introduced as follows. First, temporal dependence based methods including HA, ARIMA, SVR, and GRU achieve lower prediction performance on two real-world datasets, compared to spatio-temporal dependence based methods, i.e., T-GCN, A3T-GCN, and BSTGCN. This verifies the benefits of spatial characteristics of traffic speed for traffic prediction. Second, GCN, considering only spatial characteristics of traffic speed, is inferior to GRU in all evaluation metrics. This indicates that future traffic speed on roads is more dependent on historical traffic speed on roads. Third, we can observe that BSTGCN has a clear advantage over T-GCN and A3T-GCN, especially in the Los-loop dataset. For instance, with respect to the 15-minute prediction horizon, BSTGCN outperforms T-GCN and A3T-GCN by 1.76\% and 1.57\% in terms of $\text{R}^2$. This confirms the superior capability of BSTGCN in capturing spatio-temporal characteristics of traffic speed. Finally, we can notice that BSTGCN shows its impressive ability in mitigating long-term prediction errors. For instance, with respect to 60-minute time series, BSTGCN outperforms T-GCN and A3T-GCN by 3.85\% and 2.61\% in terms of $\text{R}^2$ in the Los-loop dataset. 
	\subsection{Visualized Results of Graph Structure}
	\begin{figure}[htbp]
		\centering
		\includegraphics[width=8.0cm]{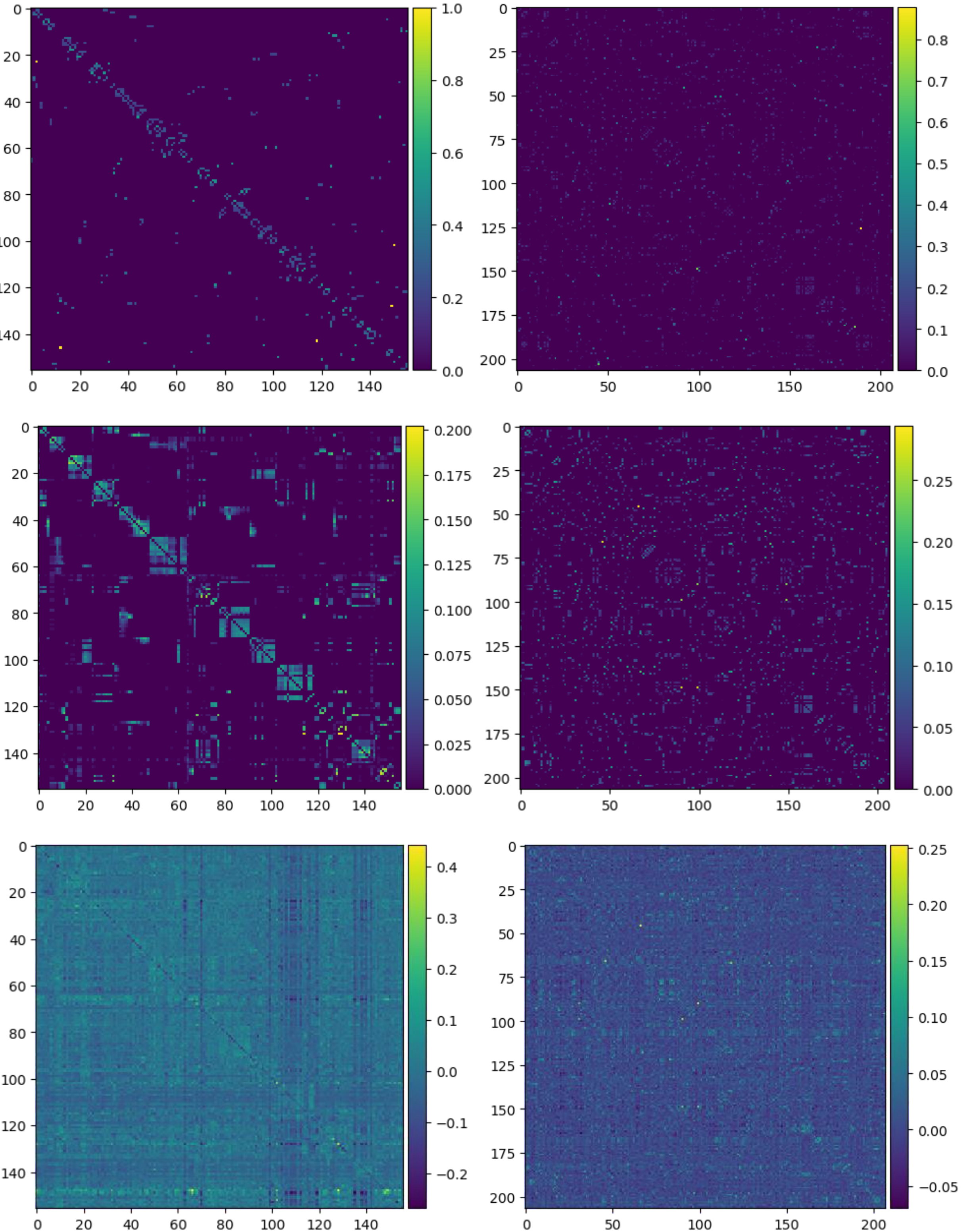}
		\caption{Visual results of graph structure. The first, second, and third row represent the observed adjacency matrix $A_{\mathcal{G}_{obs}}$, the MAP estimation of adjacency matrix $A_{\hat{g}}$, and final learned adjacency matrix $A_\mG$, respectively. The first, and second column denote the SZ-taxi and  Los-loop dataset.}
		\label{fig:A}
	\end{figure}
	\begin{figure*}[htbp]
		\centering
		\setlength{\tabcolsep}{0.2mm}
		\renewcommand{\arraystretch}{0.7} 
		\begin{tabular}{cc}
			\includegraphics[width=8.5cm]{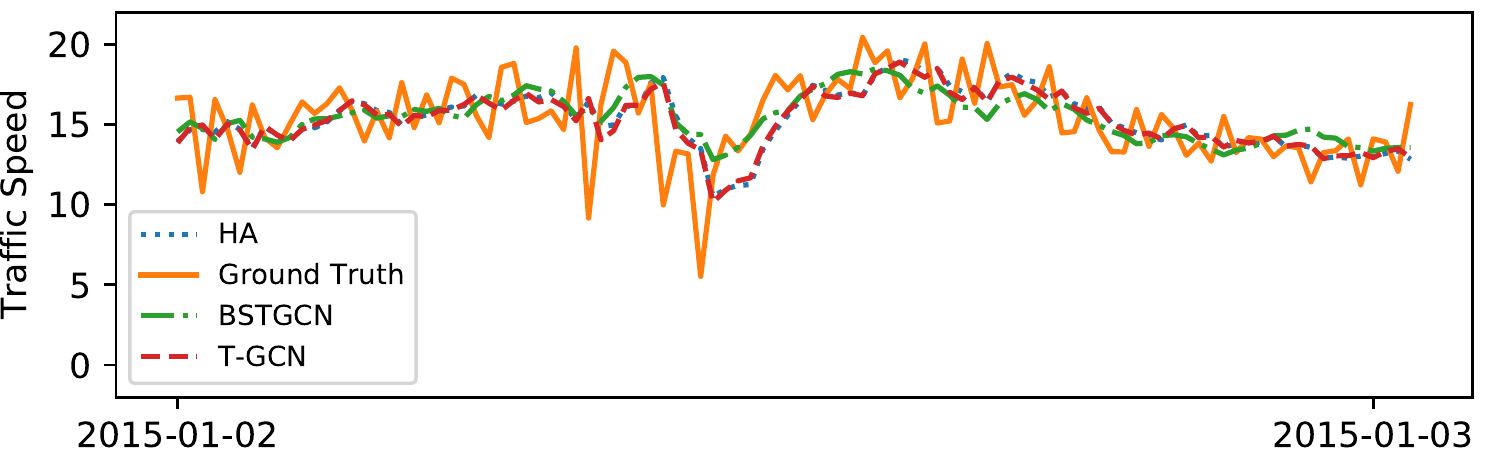}
			&\includegraphics[width=8.5cm]{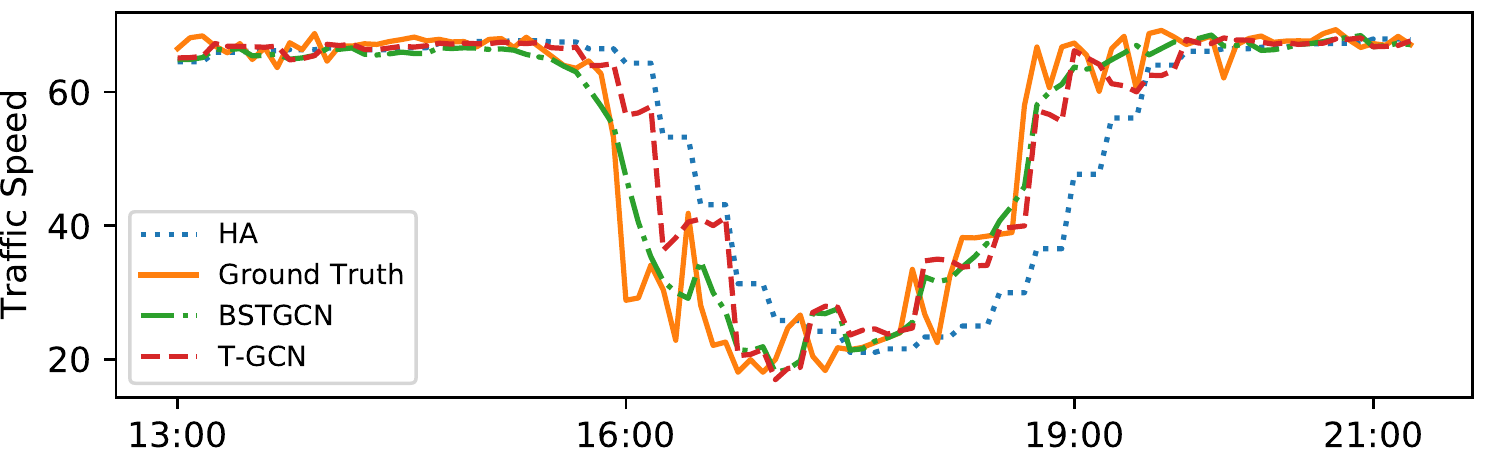}\\
			\includegraphics[width=8.5cm]{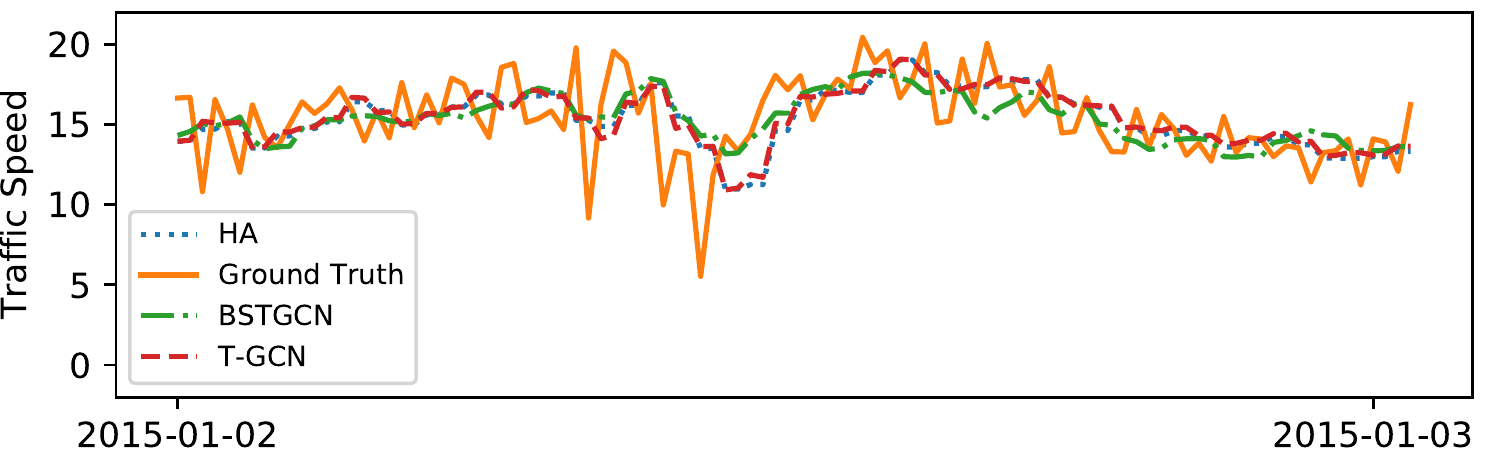}
			&\includegraphics[width=8.5cm]{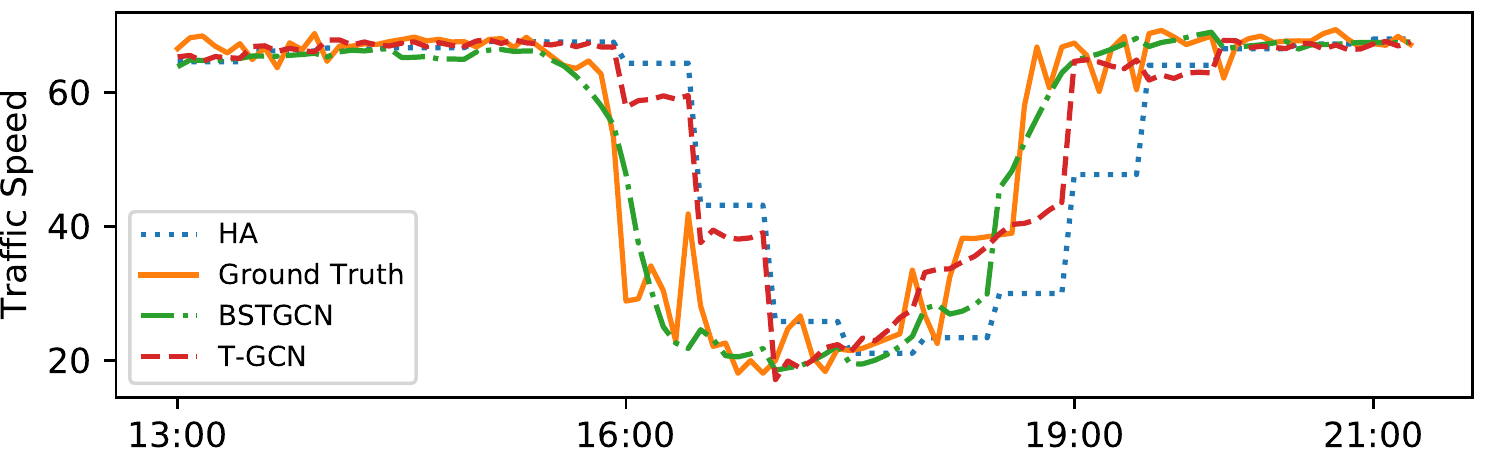}\\
			\includegraphics[width=8.5cm]{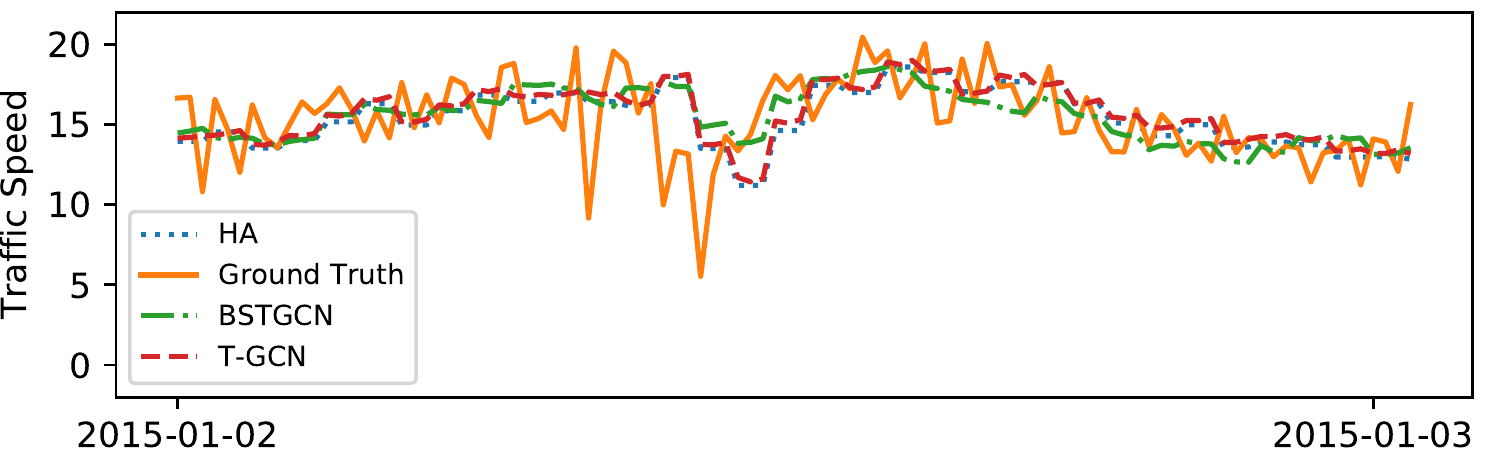}
			&\includegraphics[width=8.5cm]{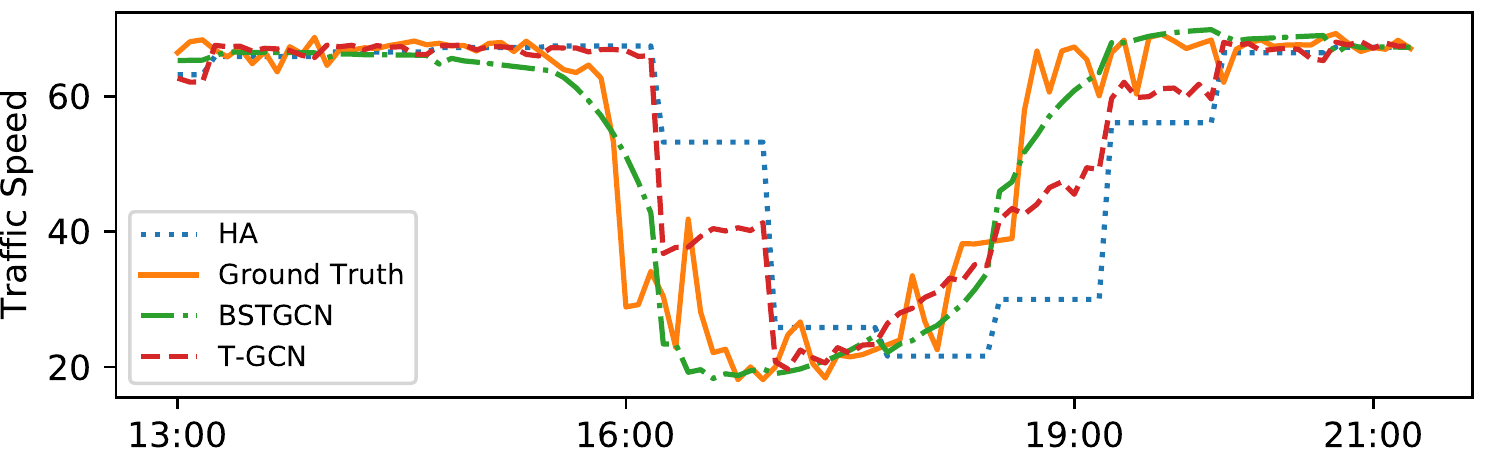}\\
		\end{tabular}
		\caption{\small Visualized prediction results of each scheme. The first, second, and third row represent the prediction horizon of 15, 30, and 45 minutes, respectively. The first, and second column denote the SZ-taxi and Los-loop dataset.}
		\label{fig:visual_comparision}
	\end{figure*}
	\par Figure~\ref{fig:A} visualizes the observed adjacency matrix $A_{\mathcal{G}_{obs}}$, the MAP estimation of adjacency matrix $A_{\hat{g}}$, and the final learned adjacency matrix $A_\mG$. We can see that $A_\mG$ has the densest connectivity among traffic flows, but is of an asymmetric structure. This confirms that the assumption that adjacency matrix is symmetric is violated in the task of traffic prediction. In addition, we can notice that the relationships among traffic flows in $A_\mG$ are not always positive. However, such negative dependence among traffic flows cannot be captured in the original observed adjacency matrix $A_{\mathcal{G}_{obs}}$. 
	\subsection{Visualized Prediction Results}
	 Figure~\ref{fig:visual_comparision} presents the prediction results of BSTGCN in the setting of 15-, 30-, and 45-minute prediction horizon. First, we can see that the traffic speed in the SZ-taxi dataset is more stationary than that of the Los-loop dataset. This explains the phenomenon that the naive historical average model (HA) performs well in the SZ-taxi dataset. Second, we can notice that the prediction results of BSTGCN are closer to the ground-truth traffic speed than that of existing methods, which is more obvious in the long-term prediction. This confirms the excellent ability of BSTGCN in traffic forecasting. Third, despite achieving impressive results, BSTGCN still cannot predict the value of sharply changing points.
	\begin{table}
		\footnotesize
		\caption{The impact of the activation function in the reset gate on the prediction performance.}
		\centering
		\resizebox{80mm}{25mm}{
			\renewcommand{\arraystretch}{1}
			\begin{tabular}{c|c|c|cc}
				\hline
				\multirow{2}{*}{Dataset}&
				\multirow{2}{*}{T}&
				\multirow{2}{*}{Metric}&
				\multicolumn{2}{c}{Activation function} \\
				\cline{4-5}
				&&&Identity&Sigmoid\\
				\hline\hline
				\multirow{5}{*}{SZ-taxi}&
				\multirow{5}{*}{60min} 
				&RMSE&4.0280&\textbf{4.0270}\\
				&&MAE&2.7020&\textbf{2.6860}\\
				&&ACC&0.7194&\textbf{0.7195}\\
				&&R$^{2}$&0.8512&\textbf{0.8513}\\
				&&VAR&\textbf{0.8516}&0.8514\\
				\hline
				\multirow{5}{*}{Los-loop}&
				\multirow{5}{*}{60min} 
				&RMSE&\textbf{6.5900}&{6.7330}\\
				&&MAE&\textbf{3.8420}&{3.9180}\\
				&&ACC&\textbf{0.8877}&{0.8853}\\
				&&R$^{2}$&\textbf{0.7766}&{0.7668}\\
				&&VAR&\textbf{0.7767}&{0.7669}\\
				\hline
		\end{tabular}}
		\label{tb:architecture}
	\end{table}
	\subsection{Deep Dive into Network Architecture}
	 We also have a deep dive into the network architecture of BSTGCN. Specifically, we switch the activation function of the reset gate in the tailored GRU from the sigmoid function to the identity function. The experimental results are presented in Table~\ref{tb:architecture}. It is interesting that such a simple modification can bring a noticeable gain in the Los-loop dataset, but has little impact in the SZ-taxi dataset. This phenomenon indicates different datasets prefer to different neural architecture. As a result, we advocate that automatically search neural architecture for the specific dataset is a promising alternative to further improve the performance of BSTGCN.
	\section{Conclusion and Future Work}
	In this paper, we propose a Bayesian spatio-temporal graph convolutional network (BSTGCN) for traffic prediction. Specifically, we propose to learn the underlying graph structure from the observed topology of the road network and traffic data in an end-to-end manner, and introduce uncertainty into the graph structure through a Bayesian approach. Experimental results on two real-world datasets verify the outstanding capability of BSTGCN in traffic prediction. In the future, we focus on two main directions. One is to extend BSTGCN to other spatio-temporal time series forecasting tasks, such as forecasting ride demand. The other is to perform neural architecture and hyperparameters search for BSTGCN.
\bibliographystyle{IEEEtran}
\bibliography{ref}
\end{document}